\pdfoutput=1

\documentclass[11pt]{article}

\usepackage[]{emnlp2021}

\usepackage{times}
\usepackage{latexsym}

\usepackage[T1]{fontenc}

\usepackage[utf8]{inputenc}

\usepackage{microtype}      
\usepackage{graphicx}       
\usepackage[T1]{fontenc}    
\usepackage{multirow}       
\usepackage{subcaption}     
\usepackage{amssymb}        
\usepackage{bold-extra}     
\usepackage{bm}             
\usepackage[hang,flushmargin]{footmisc}  
\usepackage{amsfonts,amsmath,amssymb}
\usepackage{array,booktabs}

\newcommand{\textsec}[1]{\textsection\ref{#1}}

\title{Adapted End-to-End Coreference Resolution System\\ for Anaphoric Identities in Dialogues}

\author{Liyan Xu \\
  Computer Science \\
  Emory University, Atlanta, GA \\
  \texttt{liyan.xu@emory.edu} \\\And
  Jinho D. Choi \\
  Computer Science \\
  Emory University, Atlanta, GA \\
  \texttt{jinho.choi@emory.edu} \\}

\begin{document}
\maketitle

\begin{abstract}

We present an effective system adapted from the end-to-end neural coreference resolution model, targeting on the task of anaphora resolution in dialogues.
Three aspects are specifically addressed in our approach, including the support of singletons, encoding speakers and turns throughout dialogue interactions, and knowledge transfer utilizing existing resources.
Despite the simplicity of our adaptation strategies, they are shown to bring significant impact to the final performance, with up to 27 F1 improvement over the baseline. Our final system ranks the 1st place on the leaderboard of the anaphora resolution track in the CRAC 2021 shared task, and achieves the best evaluation results on all four datasets.
\end{abstract}
\section{Introduction}
\label{sec:introduction}

Coreference resolution of anaphoric identities (a.k.a. anaphora resolution) is a long-studied Natural Language Processing (NLP) task, and is still considered one of the unsolved problems, as it demands deep semantic understanding as well as world knowledge. Although there is a significant performance boost recently by the neural decoders \citep{lee-etal-2017-end,lee-etal-2018-higher} and deep contextualized encoders such as BERT and SpanBERT \citep{joshi-etal-2019-bert,spanbert-joshi}, the majority of the experiments are based on \textit{OntoNotes} \citep{pradhan-etal-2012-conll} from the CoNLL 2012 shared task, which may overestimate the model performance due to two perspectives: the lack of support for harder cases such as singletons and split-antecedents, and the lack of focus on real-world dialogues. In this work, we target on the task of anaphora resolution in the CRAC 2021 shared task \citep{st} that addresses both perspectives, and present an effective coreference resolution system that is adapted from the recent end-to-end coreference model.

All datasets in the CRAC 2021 shared task are in the Universal Anaphora format. For simplicity, we refer to it as the UA format, and refer to the annotation scheme of the CoNLL 2012 shared task as the CoNLL format. The UA format is an extension of the CoNLL format, and further supports bridging references and discourse deixis. For anaphora resolution, the UA format differs from the CoNLL format on three aspects: the support of singletons, split-antecedents, and non-referring expressions (excluded from the current evaluation).
Our approach specifically addresses the singleton problem (Section~\ref{subsec:mr}), which is shown to be a critical component under the UA setting that brings 17-22 F1 improvement on all datasets (Section~\ref{subsec:analysis_singleton}).
Few recent work has studied the split-antecedent problem \citep{zhou-choi-2018-exist}, and we leave the split-antecedents as future work.

In addition to singletons, our approach also emphasizes on the speaker encoding (Section~\ref{subsec:speaker}) and knowledge transfer (Section~\ref{subsec:transfer}) to address the dialogue-domain perspective. Especially, we use a simple strategy of speaker-augmented encoding that captures the speaker interaction and dialogue-turn information, utilizing the strong Transformers encoder. It has been shown by the previous study that conversational metadata such as speakers can be significant for coreference resolution on dialogue documents \citep{luo-etal-2009-improving}, and we do see 2-3 F1 improvement on three datasets by simply applying the speaker encoding strategy (Section~\ref{subsec:analysis_speaker}).

Knowledge transfer from other existing resources is also shown to be important in our approach. Two different strategies are experimented, and the domain-adaptation strategy is able to bring large improvement, boosting 8 F1 for \textit{LIGHT} and 6 F1 on \textit{PSUA} (Table~\ref{tab:results}).

Our final system ranks the 1st place on the leaderboard of the anaphora resolution track in the CRAC 2021 shared task, and achieves the best evaluation results on all four datasets, with 63.96 F1 for \textit{AMI}, 80.33 F1 for \textit{LIGHT}, 78.41 F1 for \textit{PSUA}, 74.49 F1 for \textit{SWBD} (Section~\ref{subsec:results}). A brief summary of our final submission is shown in Table~\ref{tab:submission}. 

\section{Related Work}
\label{sec:related}

Pretrained Transformers encoders have been successfully adopted by recent coreference resolution models and shown significant improvement \citep{joshi-etal-2019-bert,spanbert-joshi}. We also adopt the Transformers encoder in our approach because of its superior performance.
For the neural decoder, there have been two popular directions from recent work. One is mention-ranking-based, where the model predicts only one antecedent for each mention without focusing on the cluster structure \citep{wiseman-etal-2015-learning,lee-etal-2017-end,wu-etal-2020-corefqa}. The other is cluster-based, where the model maintains the predicted clusters and performs cluster merging \citep{clark-manning-2015-entity,clark-manning-2016-improving,xia-etal-2020-incremental,yu-etal-2020-cluster}.
We adopt the mention-ranking framework in our approach because of its simplicity as well as its state-of-the-art decoding performance.

\section{Approach}
\label{sec:approach}

\subsection{Mention-Ranking (\texttt{MR})}
\label{subsec:mr}

Our baseline model \texttt{MR} adopts the mention-ranking strategy, and follows the architecture of the end-to-end neural coreference resolution model \citep{lee-etal-2017-end,lee-etal-2018-higher} with a Transformer encoder \citep{joshi-etal-2019-bert,spanbert-joshi}. Given a document with $T$ tokens, the model first enumerates all valid spans, and scores every span for being a likely mention, denoted by the mention score $s_m$. The model then greedily selects top $\lambda T$ spans by $s_m$ as mention candidates that may appear in the final coreference clusters. Let $\mathcal{X} = (x_1, \dots, x_{\lambda T})$ be the list of all mention candidates in the document, ordered by their appearance. For each mention candidate $x_i \in \mathcal{X}$, the model selects a single coreferent antecedent from all its preceding mention candidates, denoted by $\mathcal{Y}_i = (\epsilon, x_1, \dots, x_{i-1})$, with $\epsilon$ being a ``dummy'' antecedent that may be selected when $x_i$ is not anaphoric (no antecedents).

The antecedent selection is performed by the pairwise scoring process between the current mention candidate $x_i$ and each of its preceding candidate $y \in \mathcal{Y}_i$. The final pairwise score $s(x_i, y)$ consists of three scores: how likely each candidate being a mention, measured by the mention score $s_m$; and how likely they refer to the same entity, measured by the antecedent score $s_a$. The final score $s(x_i, y)$ can be denoted as follows:
\begin{align}
    s(x_i, y) = s_m(x_i) &+ s_m(y) + s_a(x_i, y, \phi(x_i,y)) \nonumber
\end{align}
Both $s_m$ and $s_a$ are computed by the FeedForward Neural Network (FFNN), and $\phi(x_i,y)$ represents additional meta features. Unlike previous work, we do not include the specific genre as a feature; instead, we simply use a binary feature on whether the document is dialogue-based or article-based, since dialogues can exhibit quite different traits from written articles \citep{aktas-stede-2020-variation}. We also adopt a speaker feature that indicates whether two candidates are from the same speaker, or whether the speaker information is not available, which is important for written articles or two-party dialogues. In Section~\ref{subsec:speaker}, we further adapt more speaker encoding to benefit multi-speaker dialogues and the personal pronoun issue.

For inference, the selected antecedent is the preceding candidate with the most pairwise score, denoted by $\text{argmax}_{y' \in \mathcal{Y}_i} s(x_i, y')$. For training, the marginal log-likelihood of all gold antecedents $\mathcal{\hat{Y}}_i \subseteq \mathcal{Y}_i$ for each $x_i \in \mathcal{X}$ is optimized, denoted by the loss $\mathcal{L}_c$:
\begin{align}
\label{eq:coref_loss}
P(y) &= \frac{e^{s(x_i, y)}}{\sum_{y' \in \mathcal{Y}_i} e^{s(x_i, y')}} \\
\mathcal{L}_c &= -\log{\prod_{x_i \in \mathcal{X}} \sum_{\hat{y} \in \mathcal{\hat{Y}}_i} P(\hat{y})}
\end{align}

\subsection{Singleton Recognition (\texttt{SR})}
\label{subsec:singleton}

\begin{figure*}[t]
\centering
\includegraphics[width=0.95\textwidth]{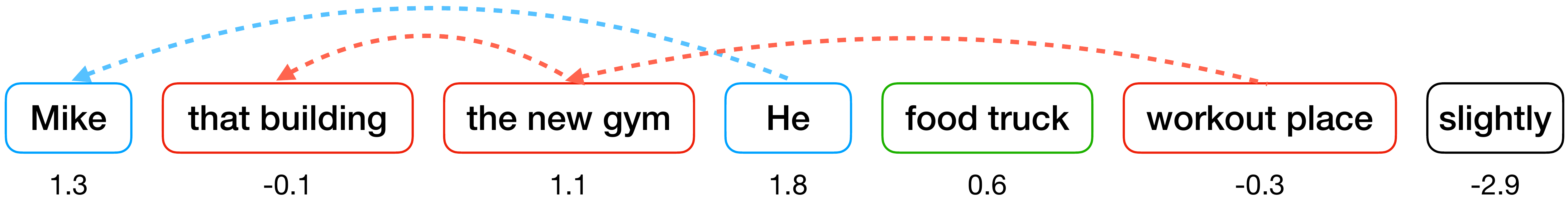}
\caption{Example of the new antecedent selection process that support singletons (Section~\ref{subsec:singleton}). Each arrow indicates the selected antecedent (the dummy antecedent is excluded), and the mention score $s_m$ is shown below each mention. Mentions of the same predicted clusters are marked in the same color. Although no antecedent is selected for ``food truck'', it will still be assigned  as a singleton cluster because of $s_m = 0.6 > 0$. ``that building'' and ``workout place'' are still assigned to the corresponding cluster even though their $s_m < 0$, to allow some slacks on the mention score prediction. ``slightly'' will not be assigned to any clusters.}
\label{fig:sr}
\vspace{-2ex}
\end{figure*}

As the UA format does support singletons, \texttt{MR} would fail to predict those singleton clusters, since the antecedent selection can only generate clusters with at least one pair of mentions. Several previous work has addressed the singleton problem from different perspectives \citep{yu-etal-2020-cluster,zaporojets2021dwie}. Our \texttt{SR} model is built upon \texttt{MR} and further recognizes singletons based on the simple strategy as follows: we make use of the mention score $s_m$ in the antecedent selection process, and create a singleton cluster for any candidates with $s_m > 0$ that have not yet found any antecedents, which poses an additional requirement on the mention score, such that only valid mentions should have $s_m > 0$. Let $\Psi^+ \subseteq \mathcal{X}$ be the set of gold mention candidates, and $\Psi^- = \mathcal{X} \setminus \Psi^+$ be the set of other mention candidates. We optimize the mention score with the binary cross-entropy loss $\mathcal{L}_m$ and joinly train with the coreference loss $\mathcal{L}_c$:
\begin{align}
    \mathcal{L}_m =& -\sum_{x_i \in \Psi^+} \log \sigma(s_m (x_i)) \nonumber \\
    & - \sum_{x_j \in \Psi^-} \log (1 - \sigma(s_m (x_j))) \label{eq:mention_optimization} \\
    \mathcal{L} =&\; \mathcal{L}_c + \alpha_m \cdot \mathcal{L}_m \label{eq:final_loss}
\end{align}
$\sigma$ is the sigmoid function, and $\alpha_m$ is a hyperparameter. $\mathcal{L}$ is the final loss composed of two tasks. In practice, we also perform negative sampling on $\Psi^-$ dynamically, so that $\Psi^+$ and $\Psi^-$ are of similar sizes ($|\Psi^+| \approx |\Psi^-|$), to alleviate the negative effects from the skewed class distribution.

In the new selection process, we still regard the selected non-dummy antecedent $y$ to be valid by $y = \text{argmax}_{y' \in \mathcal{Y}_i} s(x_i, y')$, even though the mention score of either candidate can be negative ($s_m (x_i) < 0$ or $s_m (y) < 0$). This is to allow certain slacks on the mention score prediction which could help with the mention recall. Figure~\ref{fig:sr} shows three different cases of the predicted clusters by the \texttt{SR} model.

\subsection{Speaker Encoding (\texttt{SE})}
\label{subsec:speaker}

Our \texttt{SE} model is further adapted upon \texttt{SR} model and aims to strengthen the speaker encoding for each candidate representation.
As we are targeting on the coreference resolution in dialogues, encoding speaker interactions becomes more critical, especially for the correct understanding of the speaker-grounded personal pronouns that are more frequent in dialogues than other non-dialogue genres \citep{aktas-stede-2020-variation}. 

The speaker feature introduced in Section~\ref{subsec:mr} provides shallow distinction on whether two mentions are from the same speaker. However, the speaker interactions across dialogue turns are not presented in the document encoding; therefore, the representation of each candidate has no awareness on the speaker interactions at all. To provide deeper knowledge on the interactions, we adopt a simple but effective strategy that is similar to some other work in speaker encoding \citep{le-etal-2019-speaking,wu-etal-2020-corefqa}: a special speaker token is prepended to each sentence, and we feed the new speaker-augmented document to the encoder directly.

Table~\ref{tab:se_example} shows an example on this speaker augmentation. Each speaker is indexed by the order of the first appearance in the dialogue. All the special speaker tokens are added to the tokenizer vocabulary, and will be picked up in the tokenization and encoding process. Therefore, all encoded candidate representation in the \texttt{SE} model is conditioned on the entire speaker interactions, and automatically learns to fuse the information of speakers and turns in the training process.

\begin{table}[htbp!]
\centering
\resizebox{\columnwidth}{!}{
\begin{tabular}{l}
\toprule
John: Do you know Mike? \\
Mary: He is my best friend! \\
Paul: I like him too! \\
Mary: We should meet together! \\
\midrule
\textbf{[SPK1]} Do you know Mike ? \textbf{[SPK2]} He is my best friend ! \\
\textbf{[SPK3]} I like him too ! \textbf{[SPK2]} We should meet together ! \\
\bottomrule
\end{tabular}}
\caption{Example for the speaker-augmented encoding. A special speaker token is assigned to each speaker and prepended to each corresponding sentence.}
\label{tab:se_example}
\end{table}

\subsection{Knowledge Transfer}
\label{subsec:transfer}

We also emphasize on the knowledge transfer in this task, as the training resources of dialogue corpora annotated in the UA format are limit and expensive to obtain, while there already exist larger-scale training corpora for other domains in different annotation schemes, e.g. the \textit{OntoNotes} dataset in the CoNLL format that mainly consists of non-dialogue genres. For clarity, we denote the provided data annotated in the UA format as \textit{UAD}, and other existing data in non-UA format as \textit{OD}. We investigate two common ways to make use of \textit{OD} in the training for \texttt{SE}, denoted as follows:
\begin{itemize}
    \item \texttt{SE\textsuperscript{+M}}: \textbf{M}ix \textit{OD} and \textit{UAD} together as a larger dataset, regarding \textit{OD} as data augmentation that provides more knowledge.
    \item \texttt{SE\textsuperscript{+P}}: \textbf{P}retrain the model on \textit{OD} first, then further train the model on \textit{UAD} only, regarding training on \textit{UAD} as domain adaptation.
\end{itemize}
Above two choices are plausible in our approach, because we only use data in the CoNLL format for \textit{OD}, which is still largely similar to the UA format, despite the difference on the singletons, non-referring expressions, and split-antecedents.

Similarly, we denote the model as \texttt{SR\textsuperscript{+M/P}} if the \texttt{SR} model is used instead of \texttt{SE}.

\section{Experiments}
\label{sec:experiments}

\subsection{Datasets}
\label{subsec:datasets}

For data in the UA format (\textit{UAD}), we use the ARRAU corpus \citep{arrau} from the CRAC 2018/2021 shared task. Four sub-corpora are used as the training set for \textit{UAD}, namely \textit{TRAINS-93}, \textit{PEAR}, \textit{RST}, \textit{GNOME}. One sub-corpus named \textit{TRAINS-91} is used as one of the development (dev) set. In addition, four other corpora from the CRAC 2021 shared task are also used as the development set as well as the final test set, namely \textit{AMI}, \textit{LIGHT}, \textit{Persuasion for Good} (\textit{PSUA}), \textit{Switchboard} (\textit{SWBD}). All above datasets are of the dialogue domain except for \textit{RST} and \textit{GNOME}. Table~\ref{tab:datasets} shows the detailed statistics of all \textit{UAD} datasets. Note that certain datasets do not provide speaker information, therefore their averaged numbers of speakers per document are shown as 0.

For non-UA format data (\textit{OD}), we use two datasets in the CoNLL format: \textit{OntoNotes} (\textit{ON}) \citep{pradhan-etal-2012-conll} and \textit{BOLT} \citep{li-etal-2016-large}. \textit{OntoNotes} consists of documents in six genres, where only two genres ``Telephone Conversation'' and ``Broadcast Conversation'' are of the dialogue domain; we use the same provided train/dev/test split for \textit{OntoNotes}. \textit{BOLT} has the same annotation scheme as \textit{OntoNotes} and consists of documents from discussion forums, instant messages and telephone conversations. We perform a random 80/10/10 split for the train/dev/test set of \textit{BOLT}. Detailed statistics of both datasets are shown in the bottom of Table~\ref{tab:datasets}.

\begin{table}[htbp!]
\centering
\resizebox{0.85\columnwidth}{!}{
\begin{tabular}{l||c|c|c|c}
& \tt \bf \#D & \tt \bf \#M & \tt \bf \#C & \tt \bf \#S \\
\midrule
\midrule
\it TRAINS-93 & 98 & 12148 & 4523 & 0.0 \\
\it PEAR & 20 & 3401 & 1168 & 0.0 \\
\it RST & 413 & 62409 & 38724 & 0.0 \\
\it GNOME & 5 & 5499 & 2598 & 0.0 \\
\midrule
\it TRAINS-91 & 16 & 2501 & 828 & 0.0 \\
\textit{AMI} (DEV) & 7 & 7441 & 3120 & 4.0 \\
\textit{LIGHT} (DEV) & 20 & 3448 & 1357 & 2.0 \\
\textit{PSUA} (DEV) & 21 & 2437 & 1273 & 2.0 \\
\textit{SWBD} (DEV) & 11 & 3421 & 1771 & 0.0 \\
\midrule
\textit{AMI} (TST) & 3 & 4139 & 1883 & 4.0 \\
\textit{LIGHT} (TST) & 21 & 3501 & 1359 & 2.0 \\
\textit{PSUA} (TST) & 28 & 3446 & 1857 & 2.0 \\
\textit{SWBD} (TST) & 22 & 7847 & 3897 & 2.0 \\
\midrule
\midrule
\textit{ON} (TRN) & 2802 & 155558 & 35142 & 0.6 \\
\textit{ON} (DEV) & 343 & 19155 & 4545 & 0.8 \\
\textit{ON} (TST) & 348 & 19764 & 4532 & 0.8 \\
\midrule
\textit{BOLT} (TRN) & 1110 & 58146 & 12854 & 2.5 \\
\textit{BOLT} (DEV) & 137 & 8029 & 1649 & 2.5 \\
\textit{BOLT} (TST) & 137 & 7599 & 1610 & 2.5 \\
\end{tabular}}
\caption{Statistics for all datasets (Section~\ref{subsec:datasets}), excluding non-referring expressions. \textit{UAD} datasets are shown in the upper part, and \textit{OD} datasets shown in the bottom part. TRN/DEV/TST: the train/dev/test split. \texttt{\#D}: total number of documents; \texttt{\#M}: total number of mentions; \texttt{\#C}: total number of clusters; \texttt{\#S}: averaged number of speakers per document (excluding unknown speakers).}
\label{tab:datasets}
\end{table}

\subsection{Preprocessing}
\label{subsec:preprocessing}

We only perform one trivial preprocessing step specific to the training set of \textit{UAD} datasets: remove all non-referring expressions and regard them as non-mentions, as they will not be counted in the final evaluation (Section~\ref{sec:analysis}). In addition, our current approach does not consider split-antecedents, which we will leave as future work.

\subsection{Implementation}
\label{subsec:implementation}

\begin{table*}[t!]
\begin{subtable}{\textwidth}\centering
\small
\begin{tabular}{l|ccccccccccc}
\toprule
 && \multicolumn{4}{c}{\it AMI} && \multicolumn{4}{c}{\it LIGHT} & \\
 \cmidrule{3-6} \cmidrule{8-11}
 && MUC & B\textsuperscript{3} & CEAF\textsubscript{$\phi_4$} & Avg F1 && MUC & B\textsuperscript{3} & CEAF\textsubscript{$\phi_4$} & Avg F1 \\
 \midrule
 \tt MR && 46.06 & 31.28 & 17.87 & 31.73 && 79.47 & 48.61 & 24.06 & 50.71 \\
 \midrule
 \tt SR && 54.66 & 53.32 & 53.64 & 53.87 && 79.12 & 63.34 & 65.59 & 69.35 \\
 \tt \; SR\textsuperscript{+P} && 50.08 & 52.63 & 52.60 & 51.77 && 86.39 & 74.68 & \bf 69.57 & 76.88 \\
 \midrule
 \tt SE && 57.01 & 53.91 & 53.65 & 54.86 && 77.38 & 63.91 & 65.46 & 68.92 \\
 \tt \; SE\textsuperscript{+M} && \bf 59.36 & 48.49 & 40.05 & 49.30 && 86.92 & 68.49 & 41.66 & 65.69 \\
 \tt \; SE\textsuperscript{+P} && 56.93 & \bf 55.27 & \bf 53.92 & \bf 55.37 && \bf 87.23 & \bf 74.91 & 68.86 & \bf 77.00 \\
 \midrule
 \midrule
 \tt SE\textsuperscript{+P}+DEV && 70.70 & 61.37 & 59.81 & 63.96  && 90.03 & 79.19 & 71.77 & 80.33  \\
\bottomrule
\end{tabular}
\caption{Evaluation results on the test set of \textit{AMI} and \textit{LIGHT}.}
\label{subtab:r1}
\end{subtable}
\vspace{1em}

\begin{subtable}{\textwidth}\centering
\small
\begin{tabular}{l|ccccccccccc}
\toprule
 && \multicolumn{4}{c}{\it PSUA} && \multicolumn{4}{c}{\it SWBD} & \\
 \cmidrule{3-6} \cmidrule{8-11}
 && MUC & B\textsuperscript{3} & CEAF\textsubscript{$\phi_4$} & Avg F1 && MUC & B\textsuperscript{3} & CEAF\textsubscript{$\phi_4$} & Avg F1 \\
 \midrule
 \tt MR && 75.15 & 50.80 & 21.78 & 49.24 && 72.70 & 46.65 & 22.54 & 47.30 \\
 \midrule
 \tt SR && 73.36 & 67.97 & 64.15 & 68.49 && 73.92 & 62.38 & 58.01 & 64.77 \\
 \tt \; SR\textsuperscript{+P} && 78.96 & 73.83 & 65.27 & 72.69 && 75.30 & 65.16 & 57.69 & 66.05 \\
 \midrule
 \tt SE && 72.99 & 68.56 & 64.37 & 68.64 && 74.47 & 63.32 & 58.77 & 65.52 \\
 \tt \; SE\textsuperscript{+M} && 81.63 & 65.82 & 44.83 & 64.10 && 76.54 & 61.43 & 43.97 & 60.65 \\
 \tt \; SE\textsuperscript{+P} && \bf 82.19 & \bf 76.50 & \bf 67.46 & \bf 75.38 && \bf 77.56 & \bf 67.56 & \bf 59.36 & \bf 68.16 \\
 \midrule
 \midrule
 \tt SE\textsuperscript{+P}+DEV && 84.04 & 79.57 & 71.63 & 78.41  && 80.63 & 74.39 & 68.45 & 74.49  \\
\bottomrule
\end{tabular}
\caption{Evaluation results on the test set of \textit{Persuasion for Good} (\textit{PSUA}) and \textit{Switchboard} (\textit{SWBD}).}
\label{subtab:r2}
\end{subtable}
\caption{Evaluation results on the test set of four datasets (Section~\ref{subsec:datasets}). The macro-averaged F1 of MUC, B\textsuperscript{3}, and CEAF\textsubscript{$\phi_4$} is the main evaluation metric. Section~\ref{sec:approach} describes the details of all listed approaches. \texttt{SE\textsuperscript{+P}+DEV} is the setting of our final submission to the CRAC 2021 shared task, where all available development sets are also added in the training process for \texttt{SE\textsuperscript{+P}} (Section~\ref{subsec:results}).}
\label{tab:results}
\end{table*}
\begin{table*}[htbp!]
\centering
\resizebox{0.88\textwidth}{!}{
\begin{tabular}{l|l}
\hline
\bf Track & Resolution of anaphoric identities \\
\hline
\bf Setting & Predicted mentions \\ 
\hline
\bf Baseline & \texttt{MR} (\textsec{subsec:mr}). The end-to-end coreference resolution model with the \\ 
& SpanBERT encoder \citep{spanbert-joshi,xu-choi-2020-revealing} is used as the baseline.\\
\hline
\bf Approach & \texttt{SE\textsuperscript{+P}+DEV} (\textsec{subsec:transfer}). The final model is built upon baseline with three key adaptations: \\
& 1) An updated antecedent selection process is used to support singletons, \\
& \quad with an additional optimization on the mention scores. \\
& 2) Speaker-augmentation strategy is used to encode the speakers and dialogue-turns. \\
& 3) Knowledge transfer is employed that pretrains the model on CoNLL datasets, then \\
& \quad further trains on the UA datasets as a domain adaptation step. \\
& The final submission includes the dev data into training. \\
\hline
\bf Train Data & \textit{TRAINS-93, PEAR, RST, GNOME, ON, BOLT} (\textsec{subsec:datasets}) \\
\hline
\bf Dev Data & \textit{TRAINS-91, AMI, LIGHT, PSUA, SWBD, ON, BOLT} (\textsec{subsec:datasets}) \\
\hline
\end{tabular}}
\caption{Summary of our final submission to the CRAC 2021 shared task. Train/Dev Data: all datasets we use for the training set and development set.}
\label{tab:submission}
\vspace{-0.5ex}
\end{table*}

Our system is based on the PyTorch implementation of the end-to-end coreference resolution model from \citet{xu-choi-2020-revealing}, and we follow the similar hyperparameter settings. Specifically, SpanBERT\textsubscript{Large} \citep{spanbert-joshi} is used as the Transformers encoder with maximum sequence length of 512. Long documents are split into multiple sequences, and each sequence is encoded by SpanBERT\textsubscript{Large} independently, as suggested by \citep{joshi-etal-2019-bert}. During training, we limit the maximum sequences to be 3 due to the GPU memory constraints, and a long document will be truncated into multiple documents if it exceeds the maximum sequences.

\paragraph{Hyperparameters} For all datasets, nested mentions are always enabled. We set the $\lambda = 0.5$ and maximum span width to be 30 in the span enumeration stage, and limit the maximum antecedents to be 50 in the pair scoring process. Adam optimizer is used for the optimization, with the weight decay rate of $10^{-2}$ and gradient clipping norm of $1$. We employ the learning rate of $1 \times 10^{-5}$ for Transformers parameters, and $3 \times 10^{-4}$ for task parameters. $\alpha_m = 0.1$ is used for Eq~\eqref{eq:final_loss}. In particular, we do not apply any higher-order inferences, as their benefits are shown trivial \citep{xu-choi-2020-revealing}.

\paragraph{Training} When training \textit{UAD} or \textit{OD} alone, we concatenate and mix all its corresponding corpora together as the training data. For \texttt{SE\textsuperscript{+M}}, we concatenate and mix all available training corpora together regardless of \textit{UAD} or \textit{OD}. All experiments are conducted on a Nvidia A100 GPU. 20 training epochs are used for all the settings, and the training takes around 1-2 hours for \textit{UAD} and 3-4 hours for \textit{OD}.

In particular, development sets are not added to the training data, except for our final submission to the shared task, where the best-performed model has been identified, then we train the final model with the same setting but adding all development sets in the training (Section~\ref{subsec:results}).

\section{Results and Analysis}
\label{sec:analysis}

The Universal Anaphora Scorer\footnote{\href{https://github.com/juntaoy/universal-anaphora-scorer}{https://github.com/juntaoy/universal-anaphora-scorer}} is used in the official evaluation process. For the task of anaphora resolution, the main evaluation metric is the averaged F1 score of MUC, B\textsuperscript{3} and CEAF\textsubscript{$\phi_4$}, same as the CoNLL 2012 shared task. Singletons and split-antecedents are included in the evaluation, while non-referring expressions are excluded.

\subsection{Results}
\label{subsec:results}

Table~\ref{tab:results} shows the evaluation results on the test set of four datasets using different approaches. Among all approaches without adding the dev sets into training, \texttt{SE\textsuperscript{+P}} achieves the best results on all four datasets. Another \texttt{SE\textsuperscript{+P}} model is then trained with adding the dev sets as our final submission, dentoed by \texttt{SE\textsuperscript{+P}+DEV}, which further yields the best results, and ranks the 1st place at the ``anaphoric identity'' track in the CRAC 2021 shared task.

\paragraph{Final Submission}
Table~\ref{tab:submission} lists the summary of our final submission to the shared task.

\begin{table}[t!]
\begin{subtable}{\columnwidth}\centering
\small
\begin{tabular}{l|ccccccc}
\toprule
 & \multicolumn{3}{c}{\it Mentions} && \multicolumn{3}{c}{\it Singletons} \\
 \cmidrule{2-4} \cmidrule{6-8}
 & P & R & F && P & R & F \\
 \midrule
 \tt MR & \bf 90.3 & 40.6 & 56.1 && - & - & - \\
 \midrule
 \tt SR & 84.7 & 76.4 & 80.3 && 44.7 & 54.8 & 49.2 \\
 \tt \; SR\textsuperscript{+P} & 83.1 & 74.3 & 78.5 && 42.8 & 54.0 & 47.8 \\
 \midrule
 \tt SE & 83.4 & 77.7 & 80.4 && 43.1 & \bf 55.9 & 48.7 \\
 \tt \; SE\textsuperscript{+M} & 81.5 & 69.5 & 75.0 && 29.8 & 35.7 & 32.5 \\
 \tt \; SE\textsuperscript{+P} & 84.1 & \bf 77.9 & \bf 80.9 && \bf 45.8 & 53.5 & \bf 49.4 \\
\bottomrule
\end{tabular}
\caption{Statistics on the test set of \textit{AMI}.}
\label{subtab:a1}
\end{subtable}
\vspace{1em}

\begin{subtable}{\columnwidth}\centering
\small
\begin{tabular}{l|ccccccc}
\toprule
 & \multicolumn{3}{c}{\it Mentions} && \multicolumn{3}{c}{\it Singletons} \\
 \cmidrule{2-4} \cmidrule{6-8}
 & P & R & F && P & R & F \\
 \midrule
 \tt MR & \bf 97.1 & 60.0 & 74.2 && - & - & - \\
 \midrule
 \tt SR & 87.7 & 86.8 & 87.2 && 56.3 & 72.1 & 63.2 \\
 \tt \; SR\textsuperscript{+P} & 90.1 & 89.4 & 89.7 && \bf 65.1 & 68.3 & \bf 66.6 \\
 \midrule
 \tt SE & 87.6 & 86.3 & 87.0 && 55.0 & \bf 73.6 & 62.9 \\
 \tt \; SE\textsuperscript{+M} & 91.4 & 71.7 & 80.3 && 43.1 & 23.9 & 30.8 \\
 \tt \; SE\textsuperscript{+P} & 90.0 & \bf 89.6 & \bf 89.8 && 62.1 & 68.4 & 65.1 \\
\bottomrule
\end{tabular}
\caption{Statistics on the test set of \textit{LIGHT}.}
\label{subtab:a2}
\end{subtable}
\vspace{1em}

\begin{subtable}{\columnwidth}\centering
\small
\begin{tabular}{l|ccccccc}
\toprule
 & \multicolumn{3}{c}{\it Mentions} && \multicolumn{3}{c}{\it Singletons} \\
 \cmidrule{2-4} \cmidrule{6-8}
 & P & R & F && P & R & F \\
 \midrule
 \tt MR & \bf 94.9 & 56.8 & 71.1 && - & - & - \\
 \midrule
 \tt SR & 89.5 & 85.4 & 87.4 && 66.2 & 55.4 & 60.3 \\
 \tt \; SR\textsuperscript{+P} & 91.8 & 86.4 & 89.0 && 74.3 & 51.5 & 60.8 \\
 \midrule
 \tt SE & 88.8 & 86.0 & 87.4 && 64.6 & \bf 57.2 & 60.7 \\
 \tt \; SE\textsuperscript{+M} & 90.5 & 69.9 & 78.9 && 53.6 & 26.0 & 35.1 \\
 \tt \; SE\textsuperscript{+P} & 91.9 & \bf 87.4 & \bf 89.6 && \bf 74.8 & 54.4 & \bf 63.0 \\
\bottomrule
\end{tabular}
\caption{Statistics on the test set of \textit{Persuasion for Good} (\textit{PSUA}).}
\label{subtab:a3}
\end{subtable}
\vspace{1em}

\begin{subtable}{\columnwidth}\centering
\small
\begin{tabular}{l|ccccccc}
\toprule
 & \multicolumn{3}{c}{\it Mentions} && \multicolumn{3}{c}{\it Singletons} \\
 \cmidrule{2-4} \cmidrule{6-8}
 & P & R & F && P & R & F \\
 \midrule
 \tt MR & \bf 92.0 & 54.0 & 68.1 && - & - & - \\
 \midrule
 \tt SR & 85.7 & 80.1 & 82.8 && 54.0 & 51.8 & 52.9 \\
 \tt \; SR\textsuperscript{+P} & 86.3 & 80.3 & 83.2 && 52.9 & 50.5 & 51.7 \\
 \midrule
 \tt SE & 85.0 & 80.6 & 82.7 && 53.3 & \bf 54.0 & \bf 53.7 \\
 \tt \; SE\textsuperscript{+M} & 86.3 & 68.9 & 76.6 && 39.2 & 31.7 & 35.1 \\
 \tt \; SE\textsuperscript{+P} & 87.4 & \bf 81.0 & \bf 84.1 && \bf 56.9 & 50.5 & 53.5 \\
\bottomrule
\end{tabular}
\caption{Statistics on the test set of \textit{Switchboard} (\textit{SWBD}).}
\label{subtab:a4}
\end{subtable}

\caption{Statistics of different approaches on the test set of four datasets. The left side shows the Precision/Recall/F1 (P/R/F) of the predicted mentions over gold mentions, and the right side shows the predicted singletons over gold singletons.}
\label{tab:analysis}
\end{table}

\subsection{Analysis: Singleton Recognition}
\label{subsec:analysis_singleton}

One of the main differences between the UA and CoNLL format is that UA supports singletons, as UA annotates all noun phrases. The left side of Table~\ref{tab:datasets_analysis} shows the total number and percentage of the singleton clusters on the test set of four datasets. Singletons are indeed prevalent, and all four datasets have at least 73\% of their gold clusters as singletons. Therefore, recognizing singletons can become critical for coreference resolution on the UA formatted data.

\begin{table}[htbp!]
\centering
\resizebox{\columnwidth}{!}{
\begin{tabular}{l||c|c||c|c}
& \tt \bf \#AC & \tt \bf \#SC & \tt \bf \#AM & \tt \bf \#PM \\
\midrule
\midrule
\textit{AMI} & 1883 & 1383 (73.5\%) & 4139 & 1566 (37.8\%) \\ 
\textit{LIGHT} & 1359 & 1024 (75.4\%) & 3501 & 1676 (\textbf{47.9\%})\\
\textit{PSUA} & 1857 & 1525 (\textbf{82.1\%}) & 3446 & 1464 (42.5\%) \\
\textit{SWBD} & 3897 & 2968 (76.2\%) & 7847 & 3746 (47.7\%) \\
\end{tabular}}
\caption{Statistics on the test set of all four datasets. \texttt{\#AC}: total number of all clusters. \texttt{\#SC}: total number of singleton clusters, with the corresponding percentage indicated inside parentheses. \texttt{\#AM}: total number of all mentions. \texttt{\#PM}: total number of personal pronoun mentions, with the percentage inside parentheses. All statistics exclude non-referring expressions.}
\label{tab:datasets_analysis}
\end{table}

Comparing \texttt{MR} and \texttt{SR} in Table~\ref{tab:results}, it is clear that singleton recognition plays a pivotal role in the final performance, with \texttt{SR} outperforming \texttt{MR} by a huge margin of 17-22 Avg F1 on all four datasets. To further examine the performance of \texttt{SR}, we collect the precision/recall of the predicted mentions by different models, as well as the precision/recall of predicted singletons over gold singletons, as shown in Table~\ref{tab:analysis}. Compared with \texttt{MR}, all models that support singletons receive huge gains on the mention recall with 26-36\% improvement, with relatively small 5-10\% degradation on the mention precision.

More interestingly, most \texttt{SR}/\texttt{SE}-related models are able to recover the majority of gold singletons on all four datasets, up to 73\% recall on \textit{LIGHT}, demonstrating the effectiveness of the mention score optimization in Eq~\eqref{eq:mention_optimization} and the new antecedent selection process. Nevertheless, the best F1 for singletons is still below 67 out of four datasets, suggesting that resolving singletons alone can be a challenging aspect already.

\subsection{Analysis: Speaker Encoding}
\label{subsec:analysis_speaker}

Despite the simple strategy of speaker-augmented encoding described in Section~\ref{subsec:speaker}, \texttt{SE\textsuperscript{+P}} shows decent improvement over its counterpart \texttt{SR\textsuperscript{+P}}, with 2-3\% enhancement on Avg F1 on all datasets, except for \textit{LIGHT} that has only trivial improvement, confirming that stronger speaker encoding is indeed important for the dialogue domain.

Meanwhile, \texttt{SE} does not show advantages over \texttt{SR} due to the fact that the current training corpora of all ARRAU datasets do not provide the speakers (Table~\ref{tab:datasets}); consequently, neither models could learn to use the speaker information, resulting in similar performance. This on the other side also demonstrates the significance of knowledge transfer that utilizes other existing resources.

\subsection{Analysis: Knowledge Transfer}
\label{subsec:analysis_knowledge}

Comparing the two knowledge transfer strategies, the pretraining paradigm \texttt{SE\textsuperscript{+P}} performs significantly better than the mixing paradigm \texttt{SE\textsuperscript{+M}}. In fact, while the pretraining brings improvement over \texttt{SE}, the mixing paradigm even performs worse than without knowledge transfer, likely because of the domain mismatch and the annotation format mismatch, showing that the pretraining strategy should always be preferred in this case.

The impact of the pretraining on OD can be dataset-specific, as shown by Table~\ref{tab:results}. \texttt{SE\textsuperscript{+P}} is able to boost performance upon \texttt{SE} by a good margin on \textit{AMI}/\textit{SWBD} with 0.5/2.6 F1 respectively, while \textit{LIGHT}/\textit{PSUA} can benefit significantly, with 6.7/8.1 F1 improvement. Encouraged by the results, we suggest to further explore the utilization of existing resources as a future direction.
\section{Conclusion}
\label{sec:conclusion}

In this work, we present an adapted end-to-end coreference resolution system for anaphoric identities in dialogues, specifically addressing three aspects: the support for singletons, stronger speaker and turn encoding through the dialogue interactions, as well as the knowledge transfer utilizing other existing resources. Our final system achieves the best results on all four datasets on the leaderboard of the CRAC 2021 shared task, and further analysis is performed to show the effectiveness of our proposed adaptation strategies.

\bibliography{references}
\bibliographystyle{acl_natbib}


\end{document}